# Reorganizing Educational Institutional Domain using Faceted Ontological Principles

Subhashis Das*, Debashis Naskar**, Sayon Roy***

*Dublin City University, School of computing, Glasnevin, Dublin 9, Ireland, subhashis.das@dcu.ie

**Polytechnic University of Valencia, Department of Computer Systems and Computation, (DSIC), 4 Valencia, Spain, debashis@drtc.isibang.ac.in

 ***Gangarampur College. Gangarampur, West Bengal, India, sayon@drtc.isibang.ac.in

## Abstract

The purpose of this work is to find out how different library classification systems and linguistic ontologies arrange a particular domain of interest and what are the limitations for information retrieval. We use knowledge representation techniques and languages for construction of a domain specific ontology. This ontology would help not only in problem solving, but it would demonstrate the ease with which complex queries can be handled using principles of domain ontology, thereby facilitating better information retrieval. Facet-based methodology has been used for ontology formalization for quite some time. Ontology formalization involves different steps such as, Identification of the terminology, Analysis, Synthesis, Standardization and Ordering. Firstly, for purposes of conceptualization OntoUML has been used which is a well-founded and established language for Ontology driven Conceptual Modelling. Phase transformation of "the same mode" has been subsequently obtained by OWL-DL using Protégé software. The final OWL ontology contains a total of around 232 axioms. These axioms comprise 148 logical axioms, 76 declaration axioms and 43 classes. These axioms glue together classes, properties and data types as well as a constraint. Such data clustering cannot be achieved through general use of simple classification schemes. Hence it has been observed and established that domain ontology using faceted principles provide better information retrieval with enhanced precision. This ontology should be seen not only as an alternative of the existing classification system but as a Knowledge Base (KB) system which can handle complex queries well, which is the ultimate purpose of any classification system or indexing system. In this paper, we try to understand how ontology-based information retrieval systems can prove its utility as a useful tool in the field of library science with a particular focus on the education domain.

## 1.0 Introduction

Ontologies are essential for representing knowledge in a structured way in modern knowledge-based systems. It is equally used as an important tool for information retrieval and knowledge discovery. Domain ontology helps in the formalization of linguistic knowledge through the use of logical axioms. A faceted ontology refers to an ontology divided into subtrees each of them encoding a facet, or a different aspect of domain knowledge (Prieto-Díaz 2003). It can therefore be seen as a set of classificatory knowledge-base. In this ontology, we plan to develop a faceted ontology that encodes specific domain knowledge for organizations. As per Bentivogli et al. (2004) *A domain can be defined as an area of knowledge, which is somehow recognized as unitary. A domain can be characterized by the name of a discipline where a certain knowledge area is developed (e.g. chemistry) or by the specific object of the knowledge area (e.g. space). […] Domains can be organized in hierarchies based on a relation of specificity. For instance, we can say that "tennis" is a more specific domain than "sport", or that "architecture" is more general than "town planning".*

1.1 Organization vs. Institution.

There might be confusion regarding the meaning and word-semantics of the two words "Organization" and "Institution", which are frequently and alternatively used. To take an example, there might be persistent ambivalence in efforts to semantically delineate between a medical organization and a medical institution, or for that matter between an educational organization or education institution. We try to obtain a semantic clarification of these two terms before proceeding to the next section. Oxford English Dictionary defines Organization as *"an organized body of people with a particular purpose, especially a business, society, association, etc."* and Institution *"as an organization founded for a religious, educational, professional, or social purpose."* We extend the above-quoted dictionary definition of Institution as *"a society or organization founded**,** up and running, with concrete and physical facilities associated to them for religious, educational, social, or similar purpose".*

From the above definition, it is clear that Organization is an organized body of people and it can exist without any establishment in any particular location, but Institution on the other hand should have some physical existence from which it operates and coordinates work. Educational Institution refers to an institution dedicated to education. For example, school, college, university, etc.

The educational system is not globally unanimous. It differs from country to country having varied nomenclature. We consider only the "general education system" during our classification. Other types of Education systems have been excluded, which is found, for instance in Muslim dominated countries, viz. Madrasah education (see endnote 1). We consider Bangladesh, which is a Muslim majority country. It is clear from the Bangladesh Government education system (Prodhan 2016) that they have both types of education systems, one is general and the other is Madrasah.

In general education systems, the education imparted, is common to all systems and promotes global awareness and conscious development. It embraces the traditional subjects that form and provide a shared intellectual heritage of our diverse culture. It teaches the skills of critical thinking, and of accurate and effective communication. It develops openness to the views of others, and allows revision of judgments after careful and critical thought. Such ability of comprehension is commonly and generally expected from any educated individual. General Education promotes the integration, synthesis, and application of knowledge, and includes proficiency in information literacy Cronk (2004). In this regard, UNESCO gave clear guidelines on the global education system in the International Standard Classification of Education (ISCED) 2011.

While considering the diverse codification for general systems of education available universally, existence of multiple, interchangeable, and osmotic terminology can be observed. On numerous occasions, our classification for an educational institution has alternatively been included in the concept of educational organization, by a different scheme of knowledge representation. We thus follow a faceted approach while developing our Educational Institution Ontology (EIO). Facet analysis for an existent entity has been carried out; this empowers EIO to capture and group the idea of an educational institution uniquely. Educational institutions and several related concepts are heterogeneously scattered under the broader umbrella of domain education. For a physically existing conglomerate that imparts education, there is no distinction when it comes to representing that particular conglomerate as an organization or as an institution. Among other objectives, EIO seeks to disambiguate this anomaly. EIO provides a facet based grouping that gets developed through the hierarchical tree structure. While carrying out sub-tree analysis, distinct division for educational organization and institution can be observed. Faceted-based approach has helped us to determine the division characteristics, which is a salient feature for any knowledge representation scheme.

The Rest of the paper structure is as follows: Section 2.0 provides a brief literature review on education domain and discusses the related works. Section 3.0 explains the methodology exemplifying its use for constructing the Educational Institution ontology (EIO). Section 4.0 emphasizes the core structure of the EIO Ontology. Implementation has been described in section 5.0 and model evaluation is provided in section 6.0.

## 2.0 Terminological and Ontological issues relating to the Domain of Education

Education system is diverse in nature. There might be several differences among educational systems across the continent. van Vught, Fran's, (2009) works mentions about seven types of diversity, but within the scope of our work, we have limited our discussion only to two diversities i.e. systematic diversity and structural diversity. We divided our discussion into three subcategories.

2.1 Education Domain in Library Classification

In library classification systems, the education domain is mainly categorized as a sub field under the social science category. For our discussion, we analyze categorized terminology among three major library classification systems, namely Dewey Decimal Classification (DDC), Universal Decimal Classification (UDC), and Colon Classification system (CC). The main objective is to show or verify how these classification systems treat different educational institutions within their respective scheme of classification while, accommodating the diversity.

Dewey Decimal Classification (DDC), classify the education domain under social sciences i.e. 300. Specific levels of education for instance primary education or secondary education has been classified within the range of 372-374. An important observation for this case is that, *Day care-preschool-education* in the United States of America though enumerated by this scheme, is strangely absent for all other countries. Figure 1 depicts one of such examples from DDC taken from the WebDewey browser.

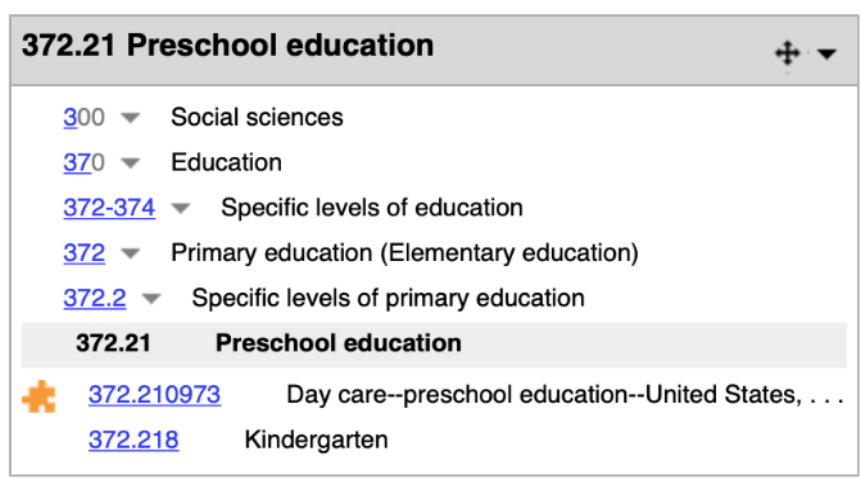

Figure 1: Classification of basic subject Education in WebDewey

DDC enunciates 'Preschool education' as an instance of 'Specific levels of primary education'. 'Preschool education' ordinally has a lower rank than 'Primary education'. Figure 2 depicts the schedule of classification of the Education domain as provided by the Colon Classification (CC) scheme. This scheme being different from enumerative schemes, uses the analytico-synthetic approach. Hence construction of schedules for different subject ideas, and inclusion of sub-concepts pertaining to subject disciplines reflect facetization and application of principles for faceted theory. In CC, the concept of 'Preschool education' ordinally has the same rank as 'Primary education'. For our purpose, we have considered the idea existent with concepts like preschool, primary, elementary education; and have designated these as 'classes'. Figure 3 illustrates, how the idea of 'Preschool education' has been accommodated in the Universal Decimal Classification (UDC) scheme. This scheme places 'Preschool education' and 'Primary education' at an ordinally similar rank. An important observation is that UDC has enumerated the concept 'Organization of preschool education' and assigned a notation (373.21).

**EDUCATION**

T [P]: [E], [2P], [2P2]

| | Foci in [P] | | | 3 | Medium of instruction | A |
|---|---|---|---|---|---|---|
| 1 | **Pre-secondary** | C | | 31 | Mother tongue | |
| 13 | Pre-school child | | | 32 | Bilingualism | |
| 15 | Elementary | | | 35 | Foreign medium | |
| | | | | 38 | Classical medium | |
| 2 | **Secondary** | | | 4 | Heuristic method | |
| 25 | Intermediate | C | | 5 | Catechism | |
| 3 | **Adult** | | | 7 | Case study. (Observation) | |
| 31 | Literate | | | 8 | Experiment | Av |
| 35 | Foreigner | | | 91 | Direct method | |
| 38 | Illiterate | | | 92 | Dramatisation | |
| | | | | 93 | Story method | |
| 4 | **University** | C | | 95 | Pictorial method | |
| 42 | Pass | | | 96 | Play method | |
| 43 | Honours | | | 97 | Lecture method | |
| 45 | Post-graduate | | | 98 | Discussion method | |
| 48 | Research | | | 4 | **Student's work** | |
| 5 | **Sex** | | | | Foci in [2P2] | |
| 51 | Male | Av | | 1 | Home work | |
| 55 | Female | | | 2 | Library work | |
| 6 | **Abnormal** | | | 3 | Study method | |
| 61 | Genius | | | 4 | Field work | |
| 62 | Idiot | | | 5 | Examination work | |
| 63 | Insane | | | 6 | Competition | |
| 65 | Criminal | | | 7 | Group work | |
| 67 | Deaf and dumb | | | 8 | Correspondence course | |
| 673 | Stammerer | | | 5 | **Educational measure-ment** | |
| 68 | Blind | | | | | |
| 7 | **Backward classes** | | | | Foci in [2P2] | |
| 9 | Other classes | | | 1 | Intelligence test | |
| | To be divided by (SD) | | | | | |
| | (Illustrative) | | | | | |

Figure 2: Basic subject Education in Colon Classification. Legend: C= class, A= Attribute and Av =Attribute value

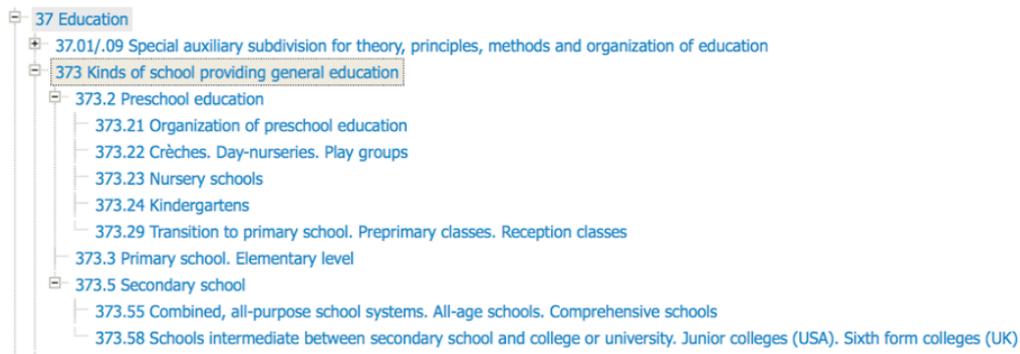

Figure 3: Classification of school in Universal Classification System (UDC)

2.2 Education Domain in Lexical Semantic Resource

The most popular lexico semantic resource is WordNet (Miller 1998). WordNet developed at Princeton University is available on an open license, since the early 1990s, has proven useful for thousands of applications to English texts. It is not flawless, but it strikes a reasonable balance between the formalization of the descriptions of lexical meaning and the wide coverage required for practical applications. In the past decade, there are many examples of WordNet developed across cultures and countries. For our investigation, we limit our discovery to Princeton WordNet (PWN) version 3.0 and all examples and flaws are taken from the same.

ISO (2011) refers to complex concepts, as those concepts that may be defined in terms of at least two other concepts. In the spirit of the faceted approach Ranganathan (1937) (as per the *meccano* property), we recommend avoiding complex concepts in the model. In fact, this would generate a concept with multiple 'is-a' parents. For instance, the concept of "red cloth" is a complex concept as it can be defined from the viewpoints of both "cloth" as well as "red". However, in this specific case "red" should be treated as a value of the attribute "color". A possible way to detect those cases is by parsing the text and to realize that "red" is actually an adjective modifying the meaning provided by the noun "cloth". ISO (pp. 40-42) describes interesting factors to consider in deciding whether to admit or not a complex concept.

Examples of complex concepts we found in Educational Institution are as follows: Day school, Night school. Day and Night modify school as they denote the timing of the school activities. In this case "day" and "night" were codified as *values of* the attribute "*school timing*".

Obsolete Concepts are those concepts whose extension does not contain entities that currently exist or that do not play the same function anymore. In other words, they are concepts that are supposedly, not to increase their extension in the future. For instance, a "chariot" is a two-wheeled horse-drawn battle vehicle used in war and races in ancient Egypt, Greece, and Rome. Therefore, on the one hand, it is unlikely that somebody produces a new one nowadays in order to use it as a battle vehicle, and on the other hand, even if it were produced it would not be used with the same purpose.

Redundant Concepts are those cases in which two or more concepts in WordNet are *equivalent*, i.e. they actually have the same meaning (see Figure 4 for example). If this is the case, the concepts must be merged as we want to avoid redundancy. In merging the two concepts, the corresponding synsets in any language need to be merged. In this case we keep all corresponding terms and generate one single gloss for it. Examples of equivalent concepts we found in Educational Institution are as follows: Kindergarten: *a preschool for children age 4 to 6 to prepare them for primary school.* Nursery school: *a small preschool for small children* was merged into Nursery school, kindergarten: *a preschool where children below the age of compulsory education, play and learn.*

A special case of redundant concepts is given when the concepts are not only (nearly) equivalent, but in a certain language, there is *polysemy*, i.e. when the concepts are lexicalized with the same term. In those circumstances, we keep the concept whose synset has a higher rank (Freihat, et al. 2013).

> School : *an educational institution*
> School : *an educational institution's faculty and students*
> were merged into
> School : *an educational institution designed for the teaching of students (or "pupils") under the direction of teachers*

Figure 4: Example of redundant concepts

Individual: There is often confusion between concepts and individuals. In natural language, individuals are typically referred to with a proper name. Concepts correspond instead to common nouns, adjectives, verbs, and adverbs. *Schema is meant to only contain concepts*, and therefore it must be free of individuals. Hence, if for some reason, during its maintenance an individual is identified, it must be deleted. For example, while refactoring Educational Institution we deleted the following: the United States military academy, United States naval academy, United states air force academy, Plato's academy

2.3 Education Domain in Formal Ontology

An ontology is a data model that represents a set of concepts within a *domain of discourse* (D') and the relationships among those concepts. Ontology is situated at the top-spectrum of the semantic hierarchy McGuinness (2002). It is used by machines to reason about the real-world objects within that domain. RDF (Resource Description Framework) is an XML-based syntax standard used for defining statements about a resource in the form of subject-predicate-object (P(S,O)) expressions called triples. RDF Schema (RDFS) defines the semantics of any particular domain with which concepts can be readily described and referred to by RDF. Miles and Bechhofer (2009) developed SKOS, "an area of work developing specifications and standards to support the use of knowledge organization systems (KOS) such as thesauri, classification schemes, subject heading systems and taxonomies within the framework of the Semantic Web". SKOS provides a standard way to represent knowledge organization systems using RDF. Encoding this information in RDF allows it to be passed between computer applications in an interoperable way. Both RDF and SKOS have limited expressive power.

OWL (Web Ontology Language) provides a more expressive ontological description of complex relations between concept pairs than RDFS does. RDFS elements can be used to define a concept in terms of a class and assigned properties in OWL.

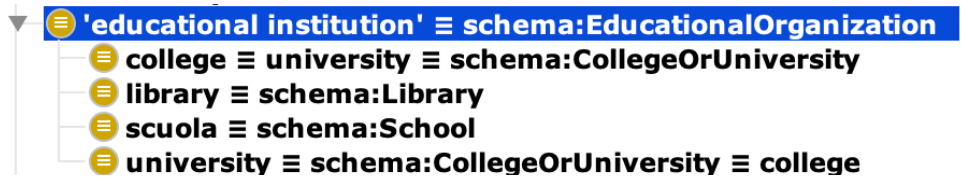

Figure 5: Classification of Educational institution in DBPedia Ontology

The purpose of building ontology can be described as some domain or as the specification of the things that make up a domain.

2.3.1 Other Related Work

This section briefly reviews the earlier literature that was produced for the education domain; like Jian Qin & Naybell Hernández (2004) constructed EduOnto, which is an Ontology for Educational Resources. In the same direction, Guangzuo et al. (2004) conducted a study on the core of OntoEdu, which is a grid-based educational system ontology for e-learning. A study constructed by authors in Ameen et al. (2012), demonstrate a method of constructing ontology in the education domain for the courses offered at universities. Interestingly, Woukeu et al. (2003) represent an ontological hypertext framework called Ontoportal for building generic web-based educational portals. The authors also describe how hypertext framework could be used to create ontology-based metadata and generate educational portals in order to obtain semantic interlinking with various web-based educational resources for learning and teaching. Educational standard is also potential for simplifying lesson planning for teachers and provides support for students by linking relevant resources. Rashid and McGuinness created education standards ontology along with a methodology for automatically generating this ontology. Rashid & McGuinness (2018). This ontology helps to improve literacy and numeracy in students and enables future potential services and creates impact for an Educational Semantic Web.

Designing integrated learning ontology for conceptualizing multilevel knowledge structures also became popular. Likewise, Chung and Kim (2016) proposed a design for syllabus integration and classification method based on the definition of the semantic model of the syllabus. To improve curricula in higher education institutions, Al-Yahya et al. (2013) present the CURONTO ontology for curriculum management. It is designed for the general management of an entire curriculum, in addition to the facilitation of program review and assessment. Similarly, a study conducted by Breis et al. (2012), where they developed a Gescur platform that supports the development of the educational curriculum and facilitates the curriculum management process. Kartis et al. (2018) purpose their work for identification and conceptualization of the entities and procedures within an academic institution and aiming to model the core concepts of a higher education curriculum. Another interesting ontological learning management system was proposed by Rani et al. (2016), where authors created an ontology to manage a learner profile so that a learner may be aligned to a proper path of learning. This ontology uses the VARK learning model which classifies what kind of learning the learner requires so that necessary resources could be provided. Despite the discussion of different models, Marcia Zeng (2008) proposed a Knowledge organization system (KOS) where the study described was based on their structures (from flat to multidimensional) and main functions. The system encompass all types of schemes for organizing information and promoting knowledge management, such as classification schemes, gazetteers, lexical databases, taxonomies, thesauri, and ontologies. On the other hand, Marcia et al. (2007) proposed an update of national and international standards (ANSI/NISO Z39.19-2005 and BS 8723) that relate to the development and encoding of KOS in the digital environment. Maria Teresa Biagetti (2020) represents the principal features of ontologies, drawing special attention to the comparison between ontologies and the different kinds of knowledge organization systems (KOS) and focuses on the semantic richness exhibited by ontologies.

Recently, the trend of using the open-source e-learning system in developing countries is more evident in comparison to the growth of the proprietary e-learning systems, this idea was by, Suteja & Jt. (2009). The transformation from traditional learning to design the e-Learning system became very popular to gain explicit knowledge. A study by Lo (2011) focuses on traditional learning and e-Learning, by adopting a value-added process orientation and an ontological instructional system. In the same way, a platform architecture for e-learning was proposed by Kaur et al. (2015), where their system consists of ontology for the e-learning process, such as teaching methods, learning styles, learning activities and course syllabus. Curriculum management and development can be improved by using ontologies in curriculum tasks like aligning, comparing, and matching between universities, educational systems or relevant disciplines. Automatically generating hypertext structures from distributed metadata is more flexible for an e-Learning system. Based on this concept, Alsultanny (2006) designs an e-Learning system by using a semantic web to provide flexible and personalized access to these learning materials. Cloud computing has received considerable attention at both, corporate firms and industry levels. Aljenaa et al. (2011) propose an e-learning framework to store rapidly developing e-learning resources on the cloud due to its scalability, thus providing E-learning as a service (EaaS). Similarly, Rani et al. (2015) proposed an ontology-driven system to implement the Felder-Silverman learning style model in addition to the learning contents, to validate its integration with the semantic web environment. Besides, cloud storage is used as the primary back-end in order to maintain the ontology. A value-adding instructional system is necessary for designing an e-Learning system. A study conducted by Lo (2011), where traditional learning and e-Learning is involved by adopting a process-oriented and systematic method of analysis. This study also draws a model called the value-added model for instructional system design for e-learning.

Nowadays, assessment has become a very important task in the course of E-Learning systems. It gives evidence of the student's intention about the courses. Kumaran & Sankar (2013), proposes a concept, using an ontology mapping, which generates a concept map based on assessments from students' learning and determining what a student knows. A similar system called Ontology E-Learning (OeLe) was implemented by Litherland et al. (2013), which automatically marks the students' free-text answers to questions of a conceptual nature. In addition, the OeLe system also provides feedback and performance evaluations to individual students and teachers. Another interesting approach was formulated by Castellanos-Nieves et al. (2011), where their methodology combines with domain ontologies, semantic annotations and semantic similarity measurements. This approach supports the assessment of open questions in eLearning courses by using Semantic Web technologies and also incorporates an algorithm for extracting knowledge from students' answers.

## 3.0 The Methodology

Our main purpose was to verify and understand all concepts associated with organizations or institutions related to the educational domain. To determine such ambiguities we choose the WordNet (Miller, 1998) as a primary resource for analysis. The organization domain of WordNet (Miller, 1998) contains about 1400 concepts. The classification is based on the definitions of the terms searched in various information sources such as the Oxford English Dictionary, United Nations (UN)'s report, Encyclopedia, Wikipedia, and other information sources. Organization is an abstract entity as per Burkhardt and Smith (1991) and we are dealing with all possible sub-divisions of organization like Business Organization, Political organization, Educational organization, Medical organization and Religious organization, etc. In this paper, we start our work by analyzing sub-tree of the "organization" which is related to education i.e. Educational Institution, and alongside keeping provisions for expansion with other sub-divisions of 'organization' in future via alignment with top-level ontology.

The following steps have been undertaken to build our formal ontological model. These steps are inspired originally from Ranganathan (1937) principle of faceted analysis as well as those provided by Giunchiglia et al. (2012, 2014), Ghosh et al. (2020) and has been combined with OntoClean methodology by Guarino & Welty (2002).

### 3.1 Information acquisition

In this step we identify relevant resources and technical documents to understand the domain of discourse.

3.1.1 Identification of relevant resources

Relevant resources has been identified by initially inspecting the terms identified during the previous step and by consulting dictionaries, available standards, and online sources (e.g. Wikipedia); the purpose is to identify the key resources necessary to deeply understand the identified terms. As a matter of fact, this requires acquiring knowledge of relevant domains. In fact, we require each single design choice to be documented by referring to an authoritative source by stressing on the advantages and disadvantages of the possible alternatives and the rationale for the selected option. For instance, in the case of educational institutions a relevant domain is education and valuable references include the International Standard Classification of Education (ISCED) provided by UNESCO (2012).

3.1.2 Study of the domain

To effectively start the analysis, it is fundamental to study the domain under examination. This allows the identification of the core terms, i.e. the terms which play a central role in the domain. For instance, for education, we referred to the ISCED standard from UNESCO aiming at a general education system. In fact, it stresses that: *"As national education systems vary in terms of structure and curricular content, it can be difficult to benchmark performance across countries over time or monitor progress towards national and international goals. In order to understand and properly interpret the inputs, processes, and outcomes of education systems from a global perspective, it is vital to ensure that data are comparable. This can be done by applying the International Standard Classification of Education (ISCED), the standard framework used to categorize and report cross-nationally comparable education statistics"*. It also provides the following classification that we mapped to the terms identified from previous steps.

3.2 Synthesis

With this step, we give shape to each facet by grouping similar concepts together. The main means for doing so is to use corresponding glosses that, according to the principle of context, should reflect the future position of the concept in the facets. In practice, this may require subsequent iterations of analysis and synthesis to progressively refine the facets and to ensure that the principles of exclusiveness, exhaustiveness, and helpful sequence are met.

3.2.1 Identification of the main characteristics of division:

In arranging identified concepts, we suggest the identification of one or more high-level characteristics of divisions which are peculiar to the domain under examination. They can be used to come up with the first two or three levels of the facet and should be reflected in the differentiation of the various analyzed terms. For instance, as it emerges from the glosses, in WordNet quite often *age* is used. junior school: *British school for children aged 7-11 and* infant school: *British school for children aged 5-7*.

However, it is well specified by Ranganathan that age is not a permanent characteristic and therefore it is not a good candidate. Among other things, the age of access to a certain level of education may vary in time and from country to country (in the above it clearly refers to the British system). Therefore, at the first level we rather distinguish educational institutions by *level of complexity* from preschool to university; at the second level we distinguished secondary schools by *programme orientation*.

3.2.2 Characteristics used

We drop "*age*" during classification as it generates time/ country bias. For example, WordNet 3.1 defined junior school as "*British school for children aged 7-11*" and infant school as "*British school for children aged 5-7*". Enrolment "age" varies from one country to another; it does not stick within a particular "age" bracket mentioned in WordNet gloss. We used "*Level of complexity*", scope and purpose of the study as deterministic characteristics for classification. The level of complexity is gradually increased from preschool to university. For choosing specific terms we follow mainly popular and most frequently used terms (i.e. Canon of currency by Ranganathan 1937). In the case of classifying higher educational institutions beyond the secondary level, we use "*Programme orientation*" as the main criteria e.g. academics, professional, and research.

3.3 Facet formulation

Facet formulation task comprise identifying classes and attributes among the identified terms selected from WordNet hierarchy.

3.3.1 Facets of entity classes

In facets of entity classes, the facet is constructed by simply looking at the genus of the definitions as it explicitly indicates the parent concept. For instance, the core structure of the facet of entity classes for the educational institution is the following:

> Educational Institution (*an institution dedicated to education*)
>   (is-a) Preschool (*an educational institution for children too young for primary school*)
>   (is-a) School (*an educational institution designed for the teaching of students under the direction of teachers*)
>     (is-a) Primary school (*a school for children where they receive the first stage of basic education*)
>     (is-a) Secondary school (*a school for students intermediate between primary school and tertiary school*)
>     (is-a) Tertiary school (*a school where programmes are largely theory based and designed to provide sufficient qualification for entry to advanced research programmes or professions with high skill requirements and leading to a degree*)
>       (is-a) Training school (*a tertiary school providing theoretical and practical training on a specific topic or leading to certain degree*)
>       (is-a) Vocational school (*a tertiary school where students are given education and training which prepares for direct entry, without further training, into specific occupation*)
>       (is-a) Technical school (*a tertiary school where students learn about technical skills required for a certain job*)
>       (is-a) Graduate school (*a tertiary school in a university or independent offering study leading to degrees beyond the bachelor's degree*)
>   (is-a) College (*an educational institution or a constituent part of a university or independent institution, providing higher education or specialized professional training*)
>   (is-a) University (*an educational institution of higher education and research which grants academic degrees in a variety of subjects and provides both undergraduate education and postgraduate education*)

Figure 6: The facet of entity class for the educational institution

### 3.3.2 Facets of attributes

fundingPolicy
    Value_of-Private (New) (New-fund collected from private bodies)
    Value_of-Public (New) (New-fund collected from government or form tax)
    Value_of- charter (New) (New-fund collected from Businessman or celebrity)

timing
    Value_of>Day (Day time of the school when it is light)
    Value_of>Night (Night time of the school between afternoon and bedtime)

facility
    Value_of> Boarding (adj)(New): the arrangement according to which pupils or students live in school during term time. The word 'boarding' is used in the sense of "bed and board," i.e., lodging and meals.

modeOfTeaching
    value_of> correspondence (school) (44786) *synonyms* Distance (New) (New- Teaching through broadcasting mode)
    value_of> regular (New) (New-Class room teaching based on schedule)

runBy
    value_of>Governmental (New-run by government body such as education department of central or state
    value_of>Religious (school) (New-run by religious body such as a church, Mosque or temple)
        -parochial
        -catholic (school)

## 4.0  Educational Institutional Ontology (EIO)

### 4.1 Core structure

In lieu of the above characteristics, we have divided educational institutions into four main sub-classes namely, preschool, school, college, and university. School is again divided into three subclasses based on three stages. While classification we consider the International Standard Classification of Education (ISCED 11) by UNESCO (2012) and we mapped it with our classification shown in the table below.

| Term and Description by UNESCO | ISCED 2011 Classification Level code | EIO Label |
|---|---|---|
| Pre-primary education | ISCED 0 | Preschool |
| Primary education | ISCED 1 | Primary school |
| Lower secondary education | ISCED 2 | Secondary school |
| Upper secondary education | ISCED 3 | Secondary school |
| Post-secondary non-tertiary education | ISCED 4 | Tertiary, Post-secondary school |
| Short-cycle tertiary education, vocationally oriented; typically prepares students to enter the workforce | ISCED 5 | Vocational school |
| Bachelor's or equivalent level | ISCED 6 | Training school, Technical school, college |
| Master's or equivalent level | ISCED 7 | Training school, Technical school, Graduate school, college |
| Advanced research programmes, Doctoral or equivalent level | ISCED 8 | University |

Table 1: Mapping between UNESCO terminology and EIO

4.2 Relationship

Hierarchical Relation. Hierarchical relation is a relation subordinate/ superordinate relationship between concepts. In the below diagram we show the hierarchical relation of the Educational Institution. It consists of *IS_A* relation.

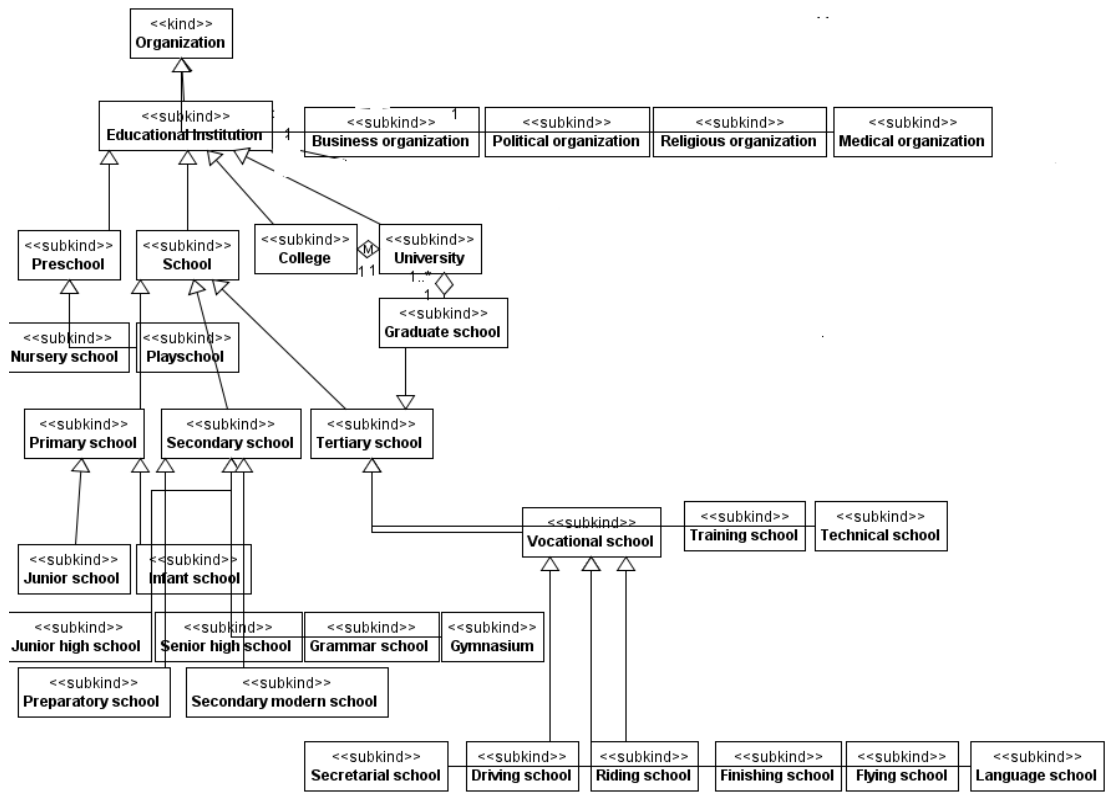

Figure 7: Hierarchy relation of the educational institution using OntoUML software

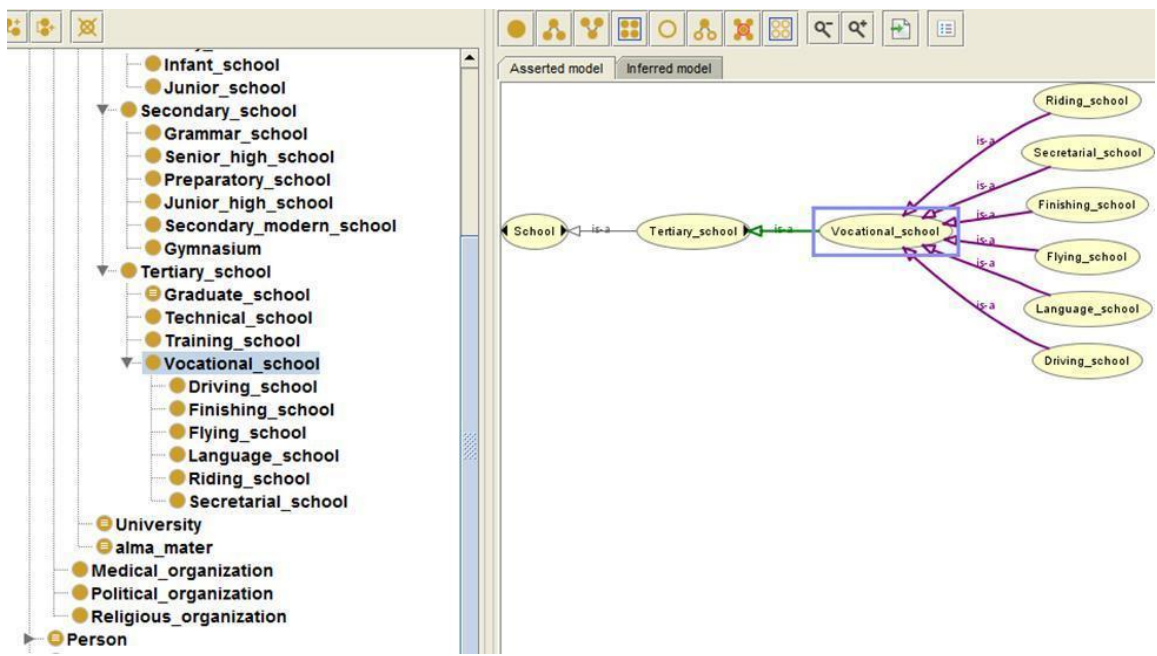

Figure 8: Hierarchy relation of the educational institution using Protégé software

Material relation. Material relation is a relation that we find in our real world. It connects two concepts with an explicit relation. In the below diagram *educational institution* (subkind) and *student* (Role) are connected with *studiedIn* relation that is derived from *enrolment* (relator). *Educational institution* (subkind) and alma mater (role) connected by *graduated* relation which is derived from *graduation* (relator). Here student is the role of a person and alma mater is the role of an educational institution.

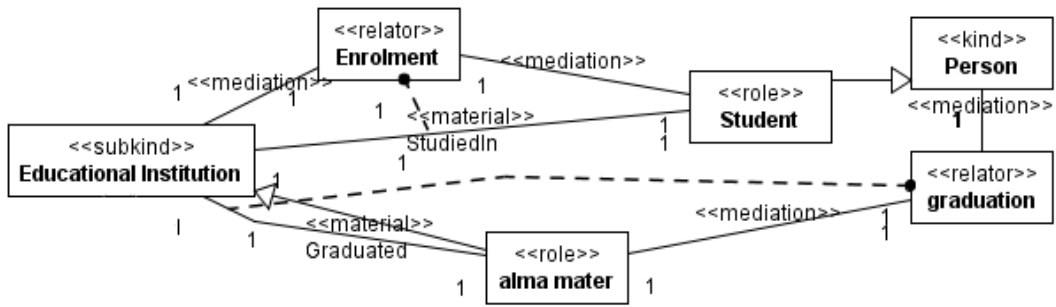

Figure 9: Material relation of enrolment in OntoUML software

Other relations are like one to many relations between college and university. In Figure 3, both *college* and *university* are the subkind of *education institutions*. But some colleges are independent and some colleges are *affiliated* under one university. With Ontological language we can easily handle this kind of situation unlike the classification system.

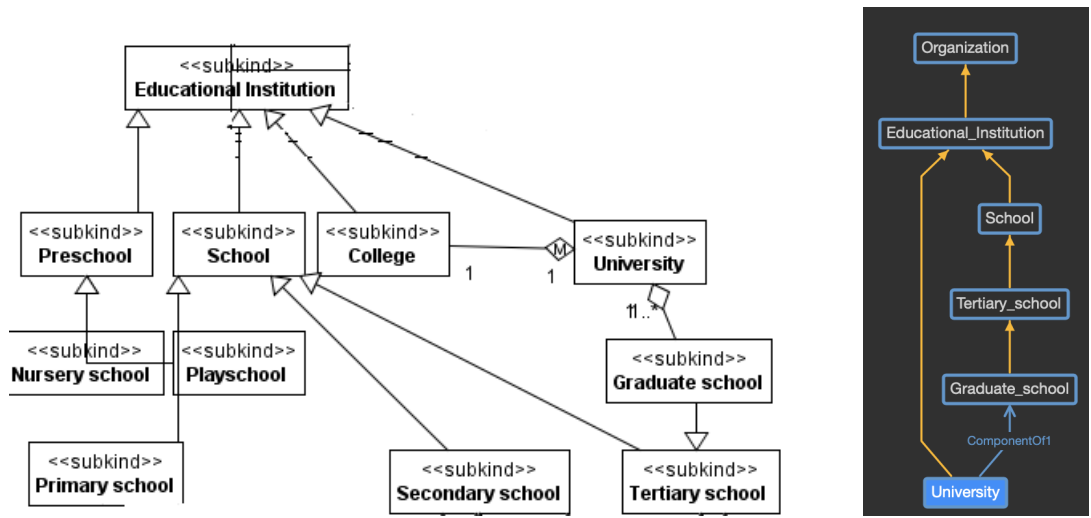

Figure 10: (Left) Class relationship of University in OntoUML software and (Right) same relationship depicted in WebProtégé user interface

<figure>

Figure 11: Details association of college and university in Protégé OWL Doc
</figure>

4.3 Process of the Work

To curate and refine our work we follow semantic approach for discerning the concept of meaning. We merged two or more concepts which has same semantic meaning and deleted those concepts from our final model which are examples for individual and not concept. More examples are provided below.

4.3.1 Merged concepts

1. School : "an educational institution faculty and student" merge with school : "an educational institution designed for the teaching of students (or "pupils") under the direction of teachers".

2. Our decision is to merge Dance school *(where students are taught to dance)* with Dancing school *(a school in which students learn to dance)* as these two are synonymous words. i.e. Only vary the word form.

3. Crammer: *"Institution that prepares pupils for an examination intensively over a short period of time"* merge with Preparatory school (New-a school generally aimed to prepare students for entry into higher educational institution) as these two are synonymous words.

4. *Kindergarten : preschool for children age 4 to 6 to prepare them for primary school"* merge with Nursery school

4.3.2 Deleted concepts
1. *Public school (45442) and Public school (45444) we deleted these two concepts from EIO as they are the attribute value. We use Public (adjective)*
2. *Private (school) (45447). We use private (adj) as it is an attribute value.*
3. *Madrasah (44814): referred to a Higher educational institution (Arabic Nation or Islamic nation); we will incorporate it in Arabic translation.*
4. *United states military academy deleted as Individual*
5. *United states naval academy as above*
6. *United states air force academy as above*
7. *Plato's academy deleted as Individual*
8. *Eton college Eton College, often informally referred to as Eton, is a British independent boarding school located in Eton delete as above*
9. *Winchester College is an independent school for boys in the British public school tradition, situated in Winchester, Hampshire, England.*
10. *Boarding school we will use adjective boarding with new gloss "the arrangement according to which pupils or students live in school during term time. The word 'boarding' is used in the sense of "bed and board," i.e., lodging and meals."*
11. *Religious (school) we will use religious i.e. attribute value*
12. *Religious school as above*
13. *catholic school same as above*
14. *church school, parochial school same as above*
15. *Day school same as above*
16. *Day (school) same as above*
17. *Night (school) same as above*
18. *charter (school) same as above*
19. *Pesantran is a Boarding school in Indonesia (space facet)*
20. *Sabbath school (a religious school providing religious education. They used to give religious education or instruction on Sunday)*

## 5.0 Implementation of the EIO model

To verify our proposed model, we test it based on data sets collected from a UK open data website. The reasons behind choosing the UK open data website are: Datasets are available in English and which meet our purpose. For the data integration task, the data is taken from the open data website data.glasgowgov.uk:

>https://data.glasgow.gov.uk/dataset/nursery-primary-secondary-and-asl-schools/resource/61370a0f-bf93-4145-8886-a7a6805b9770
>
>https://data.glasgow.gov.uk/dataset/colleges-and-universities-funded-by-scottish-funding-council/resource/b0d235cb-9f98-4d57-9d4b-8e896cf7e481

The actual datasets are csv files. The first link shows the list of Nursery, Primary, Secondary, Additional Services and Additional Support for Learning Establishments in Glasgow run by Glasgow City Council. The second CSV file shows the list of colleges and universities in Glasgow funded by Scottish Funding Council and some attributes like websites, email, city, post code and Andress. The data was integrated via conversion to excel files and the use of Protégé's cellfie plugin. An incentive for future work on the project would be to generalize our knowledge base further, to be able to integrate data from other countries into the ontology, including for example the 'istituzioni scolastiche'15 dataset from open data Trentino, the portal harvesting the metadata of Public Sector Information available on public data portals in Trentino. Working with this dataset as a basis was the initial objective of the project but has been put on hold for the time being, as there was not enough data to model the entire field of our ontology, which is why the better documented dataset of Glasgow was chosen.

## 6.0 Evaluation of the EIO model

For evaluation of the EIO model, we followed the guideline as proposed by Gómez-Pérez (1995) and Banerjee et al. (2020). According to Gómez-Pérez (1995), the goal of the evaluation process is to check what the developed ontology defined correctly, does not define, or even defines incorrectly. Two steps are needed to be followed and they are: verification and validation. The purpose of verification is to check the syntactic correctness. The purpose of validation is to check its consistency, completeness, and conciseness. Ontology editors, such as Protégé, typically provide facilities to check syntactic correctness and consistency can be checked by the reasoner such as HermiT, which are available as a Protégé' plugins. The model is complete if it fully captures what it is purported to represent of the real world. The model is concise if it does not accommodate redundancies. We ensure that the developed model is complete and concise by inducing the necessary classes and properties from the competency questions. We also use an ontology Pitfall Scanner! to check structural errors such as missing annotation, domain-range conflict as developed by Villalón et al. (2014).

One of the main advantages of ontology is to answer complex queries. In this section, we have elaborated how the EIO Ontoloɡe can be used as a knowledge-base to answer queries. Let us consider one simple query:
Who is the president of the European University Association (EUA)?

*PREFIX rdf: <http://www.w3.org/1999/02/22-rdf-syntax-ns#>PREFIX owl:*
    *<http://www.w3.org/2002/07/owl#>PREFIX xsd:*
    *<http://www.w3.org/2001/XMLSchema#>PREFIX rdfs:*
    *<http://www.w3.org/2000/01/rdf-schema#>PREFIX EI:*
    *<http://www.semanticweb.org/ontologies/2013/12/ontology.owl#> SELECT ?Person*
    *?Organization*
*WHERE { ?Person EI:PresidentOf ?Organization }*

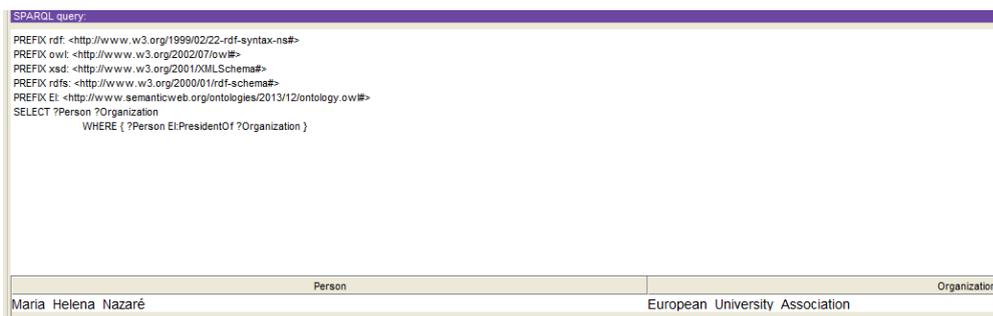

Figure 12: SPARQL query

EIO has been aligned with the top-level ontology DOLCE. It is anticipated that this will help in increasing reusability and interoperability with other ontology. The ontology is available at the web Protégé library.
https://webprotege.stanford.edu/#projects/3249bf90-bb06-4e3e-b9b7-ecf6b7bed0ab/edit/Classes

# 7.0 Conclusion and Future work

In this work we have analyzed various classification systems, and lexico-semantic resources focussing on the Education domain. Our future plan is to collect more datasets from different counties and integrate those using an open-source data integration tool. This ontology will help to guide all researchers as well as library scientists who seek information and want to explore a new way of representing library classification. We argue that our ontology-based system will provide the best results with high precision.

Endnote 1: In madrasah education, one can learn Islamic religious education along with the general education as complementary to each other in the system of education.

Acknowledgement

The authors would like to thank KnowDive group, University of Trento, Italy for providing guidance and idea of this article and to the two anonymous referees, for providing their valuable feedback. This research has received funding from the European Union's Horizon 2020 research and innovation programme under the ELITE-S Marie Skłodowska-Curie grant agreement No. 801522, by Science Foundation Ireland and co-funded by the European Regional Development Fund through the ADAPT Centre for Digital Content Technology grant number 13/RC/2106 and DAVRA Networks.